\documentclass[sigconf, nonacm, 10pt]{acmart}

\usepackage{amsmath, multirow, hyperref}

\begin{document}
\title{Multi-Type-TD-TSR - Extracting Tables from Document Images using a Multi-stage Pipeline for Table Detection and Table Structure Recognition: from OCR to Structured Table Representations}

\author{Pascal Fischer, Alen Smajic, Alexander Mehler, Giuseppe Abrami} 
\email{{s4191414@stud, s0689492@stud, amehler@em, abrami@em}.uni-frankfurt.de}
\begin{abstract}
\textit{
As global trends are shifting towards data-driven industries, the demand for automated algorithms that can convert digital images of scanned documents into machine readable information is rapidly growing. Besides the opportunity of data digitization for the application of data analytic tools, there is also a massive improvement towards automation of processes, which previously would require manual inspection of the documents. Although the introduction of optical character recognition (OCR) technologies mostly solved the task of converting human-readable characters from images into machine-readable characters, the task of extracting table semantics has been less focused on over the years. The recognition of tables consists of two main tasks, namely table detection and table structure recognition. 
Most prior work on this problem focuses on either task without offering an end-to-end solution or paying attention to real application conditions like rotated images or noise artefacts inside the document image. 
Recent work shows a clear trend towards deep learning approaches coupled with the use of transfer learning for the task of table structure recognition due to the lack of sufficiently large datasets. 
In this paper we present a multistage pipeline named \textit{Multi-Type-TD-TSR}, which offers an end-to-end solution for the problem of table recognition. It utilizes state-of-the-art deep learning models for table detection and differentiates between 3 different types of tables based on the tables' borders. For the table structure recognition we use a deterministic non-data driven algorithm, which works on all table types. We additionally present two algorithms. One for unbordered tables and one for bordered tables, which are the base of the used table structure recognition algorithm. We evaluate Multi-Type-TD-TSR on the ICDAR 2019 table structure recognition dataset \cite{8978120} and achieve a new state-of-the-art. The full source code is available on
\url{https://github.com/Psarpei/Multi-Type-TD-TSR}}.
\end{abstract}

\maketitle

\section{Introduction}
OCR based on digitized documents in general and OCR post-correction in particular remains a desideratum, especially in the context of historical documents when they have already been subjected to OCR. In the case of such texts, incorrect or incomplete recognition results often occur due to the application of mostly purely letter-oriented methods. Considerable methodological progress has been made in the recent past, with the focus of further developments being in the area of neural networks. However, special attention must be paid to the recognition of tables, where performance tends to be poor. In fact, the scores in this area are so poor that downstream NLP approaches are still practically incapable of automatically evaluating the information contained in tables. Better recognition of table structures is precisely the task to which this work relates.


\begin{figure}
  \centering
  \includegraphics[width=\linewidth]{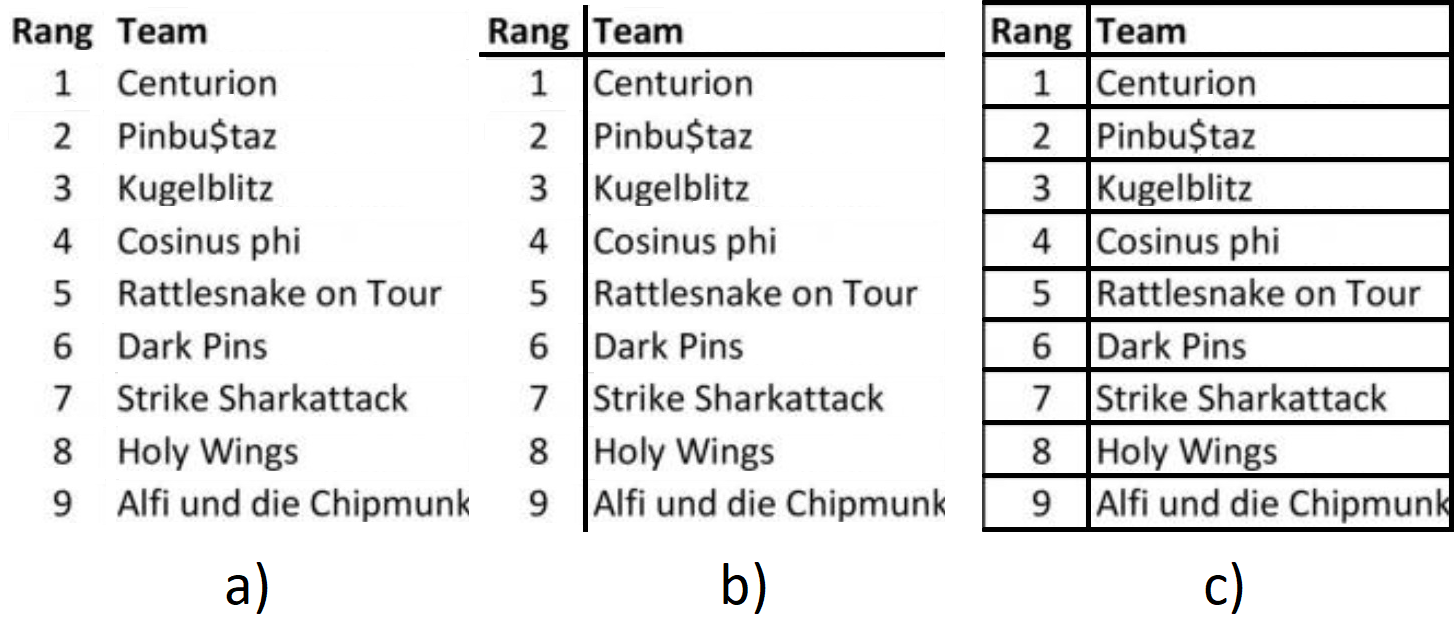}
  \caption{Types of tables based on how they utilize borders: a) tables without borders, b) tables with partial borders, c) tables with borders.}
  \label{fig:table_types}
\end{figure}

\begin{figure*}
  \centering
  \includegraphics[width=\linewidth]{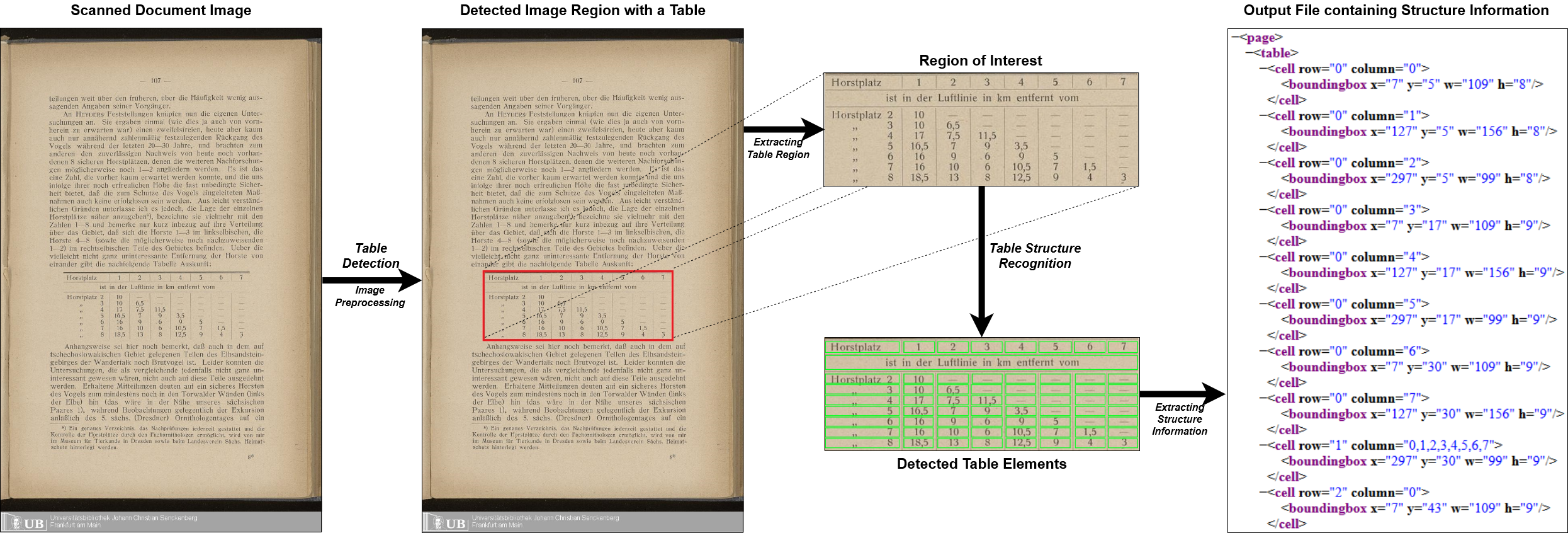}
  \caption{Schematic depiction of the staged process of \textit{Table Detection} (TD) and \textit{Table Structure Recognition} (TSR) starting from table images, i.e. digitized document pages containing tables.}
  \label{fig:extracting_semantics}
\end{figure*}

Tables are used to structure information into rows and columns to compactly visualize multidimensional relationships between information units.
In order to convert an image of a table, i.e. a scanned document that contains a table into machine readable characters, it is important to structure the information in such a way that the original relationships between the information units and their semantics is preserved. 
%
Developing algorithms that can handle such conversion tasks is a major challenge, since the appearance and layout of tables can vary widely and depends very much on the style of the table author.
%
The already mentioned row and column structure of tables goes hand in hand with different sizes and layouts of table elements, changing background colors and fonts per cell, row or column and changing borders of the table as a whole or its individual entries.
%
All these properties must be taken into account in order to achieve sufficient applicability of OCR, especially in the area of historical documents.
Otherwise information represented in tables is only insufficiently available or not available at all for downstream tasks of \textit{Natural Language Processing} (NLP) and related approaches.

Consequently, we are faced with a \textit{computer vision task} for mapping table images to structured, semantically interpreted table representations.
For this task, table borders are of particular importance because they serve as a direct visualization of the table structure and act as a frame for the elementary cell elements that are ultimately input to NLP.
In order to address this scenario, we distinguish between three types of tables.
Figure \ref{fig:table_types}a) shows a table without any table borders, Figure \ref{fig:table_types}b) a partially bordered table and Figure \ref{fig:table_types}c) a fully bordered one. 
We refer to tables without any borders as \textit{unbordered tables}, tables that are completely bordered as \textit{bordered tables} and tables that contain some borders as \textit{partially bordered tables}. It should be mentioned that \textit{partially bordered tables} also include unbordered and bordered tables.

The task of converting an image of a table into machine readable information starts with the digital image of a document, which is created using a scanning device. 
Obviously, this process is crucial for the later conversion, since small rotations of the document during scanning or noise artifacts generated by the scanning device can have a negative impact on recognition performance. 
The conversion itself involves two steps, namely \textit{Table Detection} (TD) inside a document image and \textit{Table Structure Recognition} (TSR).
%
TD is performed to identify all regions in images that contain tables, while TSR involves identifying their components, i.e. rows, columns, and cells, to finally identify the entire table structure - see Figure \ref{fig:extracting_semantics} for a schematic depiction of this two-step recognition process.
%
In general, these two tasks involve detecting some sort of bounding boxes that are correctly aligned with the table elements to be identified.
%
However, without proper alignment of the entire table image, it is not possible to generate accurate bounding boxes, which reduces the overall performance of table representation. 
Thus, the correct alignment of table images is to be considered as a constitutive step of the computer vision task addressed here.

\begin{figure*}
  \centering
  \includegraphics[width=\linewidth]{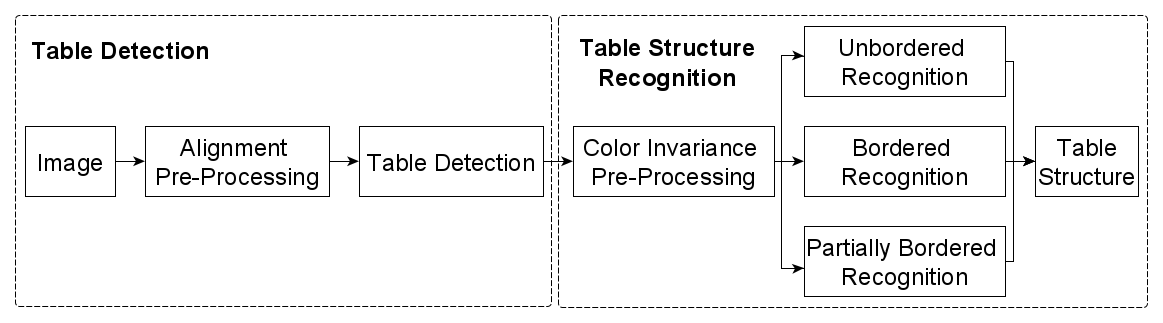}
  \caption{The two-stage process of TD and TSR in Multi-Type-TD-TSR.} 
  \label{fig:multi-stage_pipeline}
\end{figure*}

In this paper, we present a multistage pipeline named \textit{Multi-Type-TD-TSR} which solves the task of extracting tables from table images and representing their structure in an end-to-end fashion.
%
The pipeline consists of four main modules that have been developed independently and can therefore be further developed in a modular fashion in future work. 
%
Unlike related approaches, our pipeline starts by addressing issues of rotation and noise that are common when scanning documents.
For TD we use a fully data-driven approach based on a \textit{Convolutional Neural Network} (CNN) to localize tables inside images and forward them to TSR. 
For the latter we use a deterministic algorithm in order to address all three table types of Figure \ref{fig:table_types}. 
%
Between TD and TSR, we perform a pre-processing step to create font and background color invariance so that the tables contain only black as the font color and white as the background color.
%
In addition, we present two specialized algorithms for bordered tables and unbordered tables, respectively.
%
These two algorithms form the basis for our table type-independent algorithm for recognizing table structures.
%
%
Our algorithm finally represents recognized table elements by means of a predefined data structure, so that the table semantics can finally be processed further as a structured document -- for example in NLP pipelines.

%
For table detection we use the state-of-the-art deep learning approach proposed by Li \textit{et al.} \cite{li-etal-2020-tablebank}, which was evaluated on the TableBank  \cite{li-etal-2020-tablebank} dataset.
%
For the task of table structure recognition we evaluate 
Multi-type-TD-TSR on the ICDAR 2019 dataset (Track B2) \cite{8978120}.

The paper is structured as follows: In Section \ref{relatedwork}, we summarize related work. 
Then the pipeline of Multi-type-TD-TSR is explained in Section \ref{fig:multi-stage_pipeline} and in more detail in Section \ref{methods}. 
After that, the evaluation and comparison with state-of-the-art techniques is presented in Section \ref{evaluation}. 
Finally, we draw a conclusion in Section \ref{conclusion} and preview future work in Section \ref{futurework}.

\section{Related Work}
\label{relatedwork}
Several works have been published on the topic of extracting table semantics and there are comprehensive surveys available describing and summarizing the state-of-the-art in the field since 1997 when P. Pyreddy and, W. B. Croft \cite{pyreddi1997system} proposed the first approach of detecting tables using heuristics like character alignment and gaps inside table images. To further improve accuracy, Wonkyo Seo \textit{et al.} \cite{seo2015junction} used an algorithm to detect junctions, that is, intersections of horizontal and vertical lines inside bordered tables. 
T. Kasar \textit{et al.} \cite{kasar2013learning} also used junction detection, but instead of heuristics he passed the junction information to a SVM \cite{cortes1995support} to achieve higher detection performance.

With the advent of \textit{Deep Learning} (DL), advanced object recognition algorithms, and the first publicly available datasets, the number of fully data-driven approaches continued to grow. 
Azka Gilani \textit{et al.} \cite{gilani2017table} was the first to propose a DL-based approach for table detection by using Faster R-CNN \cite{ren2015faster}. 
Currently, there are three primary datasets used for TD and TSR. 
The first one is provided by the ICDAR 2013 table competition \cite{6628853}, which is a benchmark for TD and TSR. 
The dataset contains a total of 150 tables: 75 tables in 27 excerpts from the EU and 75 tables in 40 excerpts from the US Government. Its successor, the ICDAR 2019 competition on Table Detection and Recognition (cTDaR) \cite{8978120} features two datasets. The first one consists of modern documents with modern tables, while the other consists of archival documents with presence of hand-drawn tables and handwritten text. In general this dataset includes 600 documents for each of the two datasets with annotated bounding boxes for the image regions containing a table. In 2020, Li \textit{et al.} \cite{li-etal-2020-tablebank} published the TableBank dataset, the latest available dataset. 
It consists of Word and LaTeX documents and is the first to be generated completely automatically. 
The TableBank dataset includes 417.234 annotated tables for TD and 145.463 for TSR. 
%
Unfortunately, the dataset for TSR contains only information about the number of rows and columns, but no information about the location or size of table elements.
Li \textit{et al.} \cite{li-etal-2020-tablebank} proposed a ResNeXt-152 \cite{xie2017aggregated} model trained on TableBank for TD, which represents the current state-of-the-art for type-independent TD.

In 1998, Kieninger and Dengel \cite{kieninger1998t}, introduced the first approach to TSR by means of clubbing text into chunks and dividing chunks into cells based on column border. 
The low number of annotated tables with bounding-boxes for TSR enforces the use of transfer learning and leads to a high risk of overfitting. 
Schreiber \textit{et al.} \cite{schreiber2017deepdesrt} addressed TD and TSR in a single approach using a two-fold system based on Faster R-CNN \cite{ren2015faster} for TD and DL-based semantic segmentation
for TSR that utilizes transfer learning. 
Mohsin \textit{et al.} \cite{reza2019table} generalized the model by combining a GAN \cite{goodfellow2014generative} based architecture for TD with a SegNet \cite{badrinarayanan2017segnet} based encoder-decoder architecture for TSR. 
Recently, Prasad \textit{et al.} \cite{prasad2020cascadetabnet} presented an end-to-end TD and TSR model based on transfer learning.
It is the first approach that distinguishes between different types of tables and solves each type with a different algorithm. 
First, an object detection model is used to identify tables inside a document and to classify them as either bordered or unbordered. 
In a second step, TSR is addressed with a CNN for non-bordered tables and with a deterministic algorithm based on the vertical and horizontal table borders for bordered tables. 
For the development of Multi-Type-TD-TSR we utilize the TD approach proposed by Li \textit{et al.} \cite{li-etal-2020-tablebank}, since it offers the state-of-the-art model for type-independent TD, which is crucial for our TSR approach. For the task of TSR we use the architecture proposed by Prasad \textit{et al.} \cite{prasad2020cascadetabnet}, who introduced the erosion and dilation operation for bordered tables, and extend this approach to implement a robust algorithm that can handle all table types.

\section{End-to-End Multistage pipeline}
\label{multistagepipeline}
Figure \ref{fig:multi-stage_pipeline} shows our multi-stage, multi-type pipeline. 
It consists of two main parts, namely \textit{Table Detection} (TD), which processes the full-size image, and \textit{Table Structure Recognition} (TSR), which processes only the recognized sections from TD. 
In the first step, a pre-processing function is applied to the scanned document image to correct the image alignment before passing the scan to the next step.
The aligned image is then fed into into a ResNext152 model of the sort proposed by \cite{li-etal-2020-tablebank} in order to perform TD.
Based on the predicted bounding boxes, the recognized tables are cropped from the image and successively passed to TSR. 
%
Here, our algorithm applies a second preprocessing step that converts the foreground (lines and fonts) to black and the background to white to create a color-invariant image. 
%
In the next step, 3 branching options are available.
The first one uses an algorithm specialized for unbordered tables. 
The second one utilizes a conventional algorithm based on \cite{prasad2020cascadetabnet} that is specialized for bordered tables. 
%
The third option is a combination of the latter two; it works on partially bordered tables, which includes fully bordered and fully unbordered tables, making the algorithm type-independent.
%
Finally, the recognized table structure is exported per input table using a predefined data structure.

\begin{figure}
  \centering
  \includegraphics[width=\linewidth]{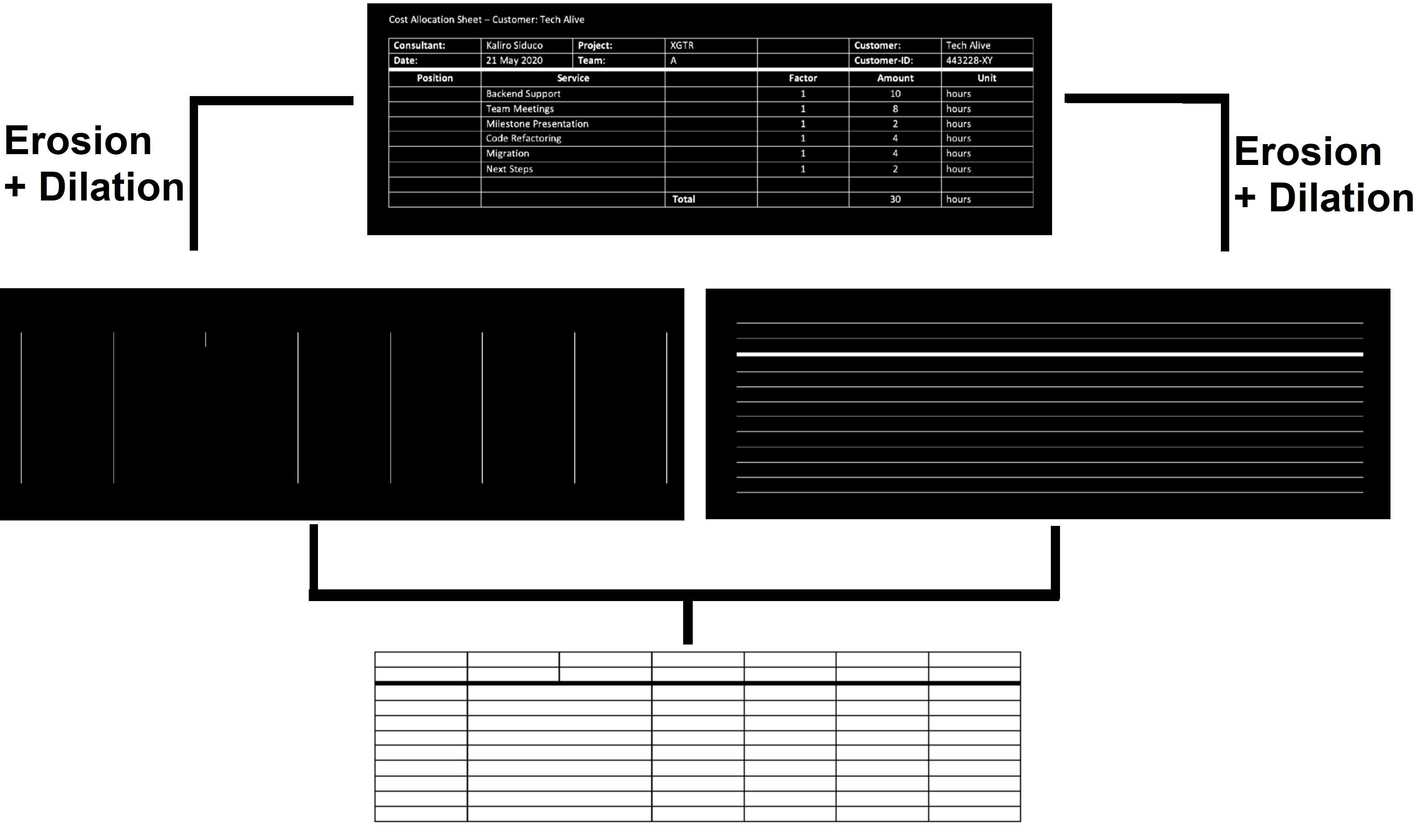}
  \caption{Example of the erosion and dilation operations as performed by Multi-Type-TD-TSR for bordered tables.}
  \label{fig:bordered_algo}
\end{figure}

\section{Methods}
\label{methods}
All three TSR algorithms used by us are based on the following two mathematical operations.
The first operation is \texttt{dilation} defined as follows:
\begin{equation}
    \texttt{dilation} (x,y) = \max _{(x',y'): \, \texttt{K} (x',y') \ne0 } I (x+x',y+y')
\end{equation} 
This operation involves filtering an image $I$ with a kernel $K$, which can be of any shape and size, but in our case is a rectangle. 
$K$ has a defined anchor point, in our case the center of the kernel.
%
As $K$ slides over the image, the maximum pixel value overlapped by $K$ is determined and the image pixel value at the anchor point position is overwritten by this maximum value. 
%
Maximization causes bright areas within an image to be magnified.

The second operation named \texttt{erosion} is defined as follows:
\begin{equation}
    \texttt{erosion} (x,y) = \min _{(x',y'): \, \texttt{K} (x',y') \ne0 } I (x+x',y+y')
\end{equation}
It works analogously to \texttt{dilation}, but determines a local minimum over the area of the kernel. 
%
As $K$ slides over $I$, it determines the minimum pixel value overlapped by $K$ and replaces the pixel value under the anchor point with this minimum value. 
Conversely to \texttt{dilation}, \texttt{erosion} causes bright areas of the image to become thinner while the dark areas become larger.

Following the example of Prasad \textit{et al.} \cite{prasad2020cascadetabnet}, we use \texttt{erosion} on bordered tables to detect vertical and horizontal borders, which need to be retained, while removing the font and characters from the table cells resulting in a grid-cell image. \texttt{Dilation} is applied successively to restore the original table border structure, since \texttt{erosion} shortens the borders. Additionally we apply \texttt{erosion} on unbordered tables to add the missing borders producing a full grid-cell image.

\subsection{Table Alignment Pre-Processing}
The first method of our Multi-Type-TD-TSR algorithm includes table alignment pre-processing, which is crucial for TSR. 
Currently, this pre-processing utilizes the text skew correction algorithm proposed by \cite{pyimageresearch}. 
To remove all noise artifacts within an image, we apply a median filter of kernel size 5x5 pixels, which showed the best results in our experiments. 
%
One by one, the algorithm converts the image to grayscale and flips the pixel values of the foreground (lines and characters) so that they have a white color and the background has a black color. 
In the next step we compute a single bounding box that includes all pixels that are not of black color and therefore represent the document content. 
Based on this bounding box we calculate its rotation and apply a deskew function, which rotates the bounding box along with its content to be properly aligned.

\subsection{Table Detection}
In the TD step, we extract the bounding-boxes for each table inside the image by using a \textit{Convolutional Neural Network} (CNN) which does not distinguish between the three table types (see Figure \ref{fig:table_types}).
We utilize the approach of Li \textit{et al.} \cite{li-etal-2020-tablebank} who trained a ResNeXt-152 \cite{xie2017aggregated} model on the TableBank dataset \cite{li-etal-2020-tablebank}. 
The reason for this selection is that this model reaches the best results by only detecting bounding boxes for each table without classification. The state-of-the-art approach from Prasad \textit{et al.} performs an additional classification of tables by borders. 
We decided against the approach of Prasad \textit{et al.} \cite{prasad2020cascadetabnet}, since their classification only considers two table types and also includes a slightly different definition of bordered and unbordered tables than ours.

\begin{figure}
  \centering
  \includegraphics[width=\linewidth]{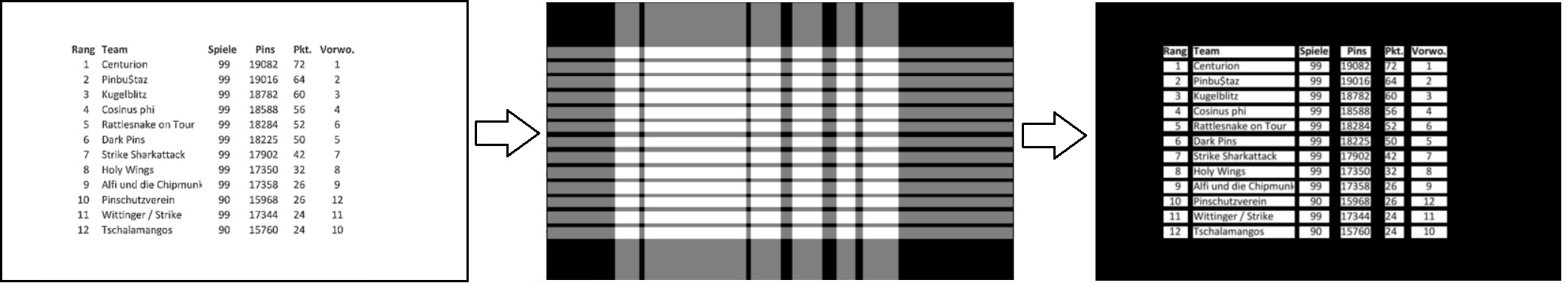}
  \caption{Example of the erosion operation as performed by  Multi-Type-TD-TSR for unbordered tables.}
  \label{fig:unbordered_algo}
\end{figure}
\subsection{Bordered TSR}
The algorithm for bordered TSR is based on the same named algorithm from Prasad \textit{et al.} \cite{prasad2020cascadetabnet}, which utilizes the erosion and dilation operation for extracting the row-column grid cell image without any text or characters. 
The first step includes converting the image into a binary representation with pixel values of either zero (black) or one (white) and finally inverting these values to get a table image of white foreground (lines and characters) and black background as shown in the upper part of Figure \ref{fig:bordered_algo}. 
%
In the next step, a horizontal and vertical erosion kernel $k_h, k_v \in \mathbb{R}^2$ are applied independently to the inverted image.
%
It is worth mentioning that the kernel shape and size is not fixed and can be set for both erosion kernels. 
%
The erosion kernels are generally thin vertical and horizontal strips that are longer than the overall font size but shorter than the size of the smallest grid cell and, in particular, must not be wider than the smallest table border width. 
%
Using these kernel size constraints results in the erosion operation removing all fonts and characters from the table while preserving the table borders.
Since the erosion operation is keeping the minimum pixel value from the kernel overlay, its application leads to shorter lines compared to the original table borders. In order to restore the original line shape, the algorithm applies the dilation operation using the same kernel size on each of the two eroded images like shown in the middle part of Figure \ref{fig:bordered_algo}, producing an image with vertical and a second with horizontal lines. The dilation operation rebuilds the lines by keeping only the maximum pixel value from the kernel overlay of the image. 
%
Finally, the algorithm combines both images by using a \textit{bit-wise or} operation and re-inverting the pixel values to obtain a raster cell image, as shown in the lower part of Figure \ref{fig:bordered_algo}.
We then use the contours function on the grid-cell image to extract the bounding-boxes for every single grid cell.

\subsection{Unbordered TSR}

The TSR algorithm for unbordered tables works similarly to the one for bordered tables. 
It also starts with converting the image to a binary representation. However, unlike the first algorithm it does not invert the pixel values straight away and also does not utilize the dilation operation. 
Furthermore, it uses a different kind of erosion compared to TSR for bordered tables. 
The erosion kernel is in general a thin strip with the difference that the horizontal size of the horizontal kernel includes the full image width and the vertical size of the vertical kernel the full image height. 
The algorithm slides both kernels independently over the whole image from left to right for the vertical kernel, and from top to bottom for the horizontal kernel. During this process it is looking for empty rows and columns that do not contain any characters or font. The resulting images are inverted and combined by a \textit{bit-wise and} operation producing the final output as shown in the middle part of Figure \ref{fig:unbordered_algo}. This final output is a grid-cell image similar to the one from TSR for bordered tables, where the overlapping areas of the two resulting images represent the bounding-boxes for every single grid cell as shown in the right part of Figure \ref{fig:unbordered_algo} which displays the grid cells produced by our TSR algorithm and the corresponding text.

\subsection{Partially Bordered TSR}
\begin{figure}
  \centering
  \includegraphics[width=\linewidth]{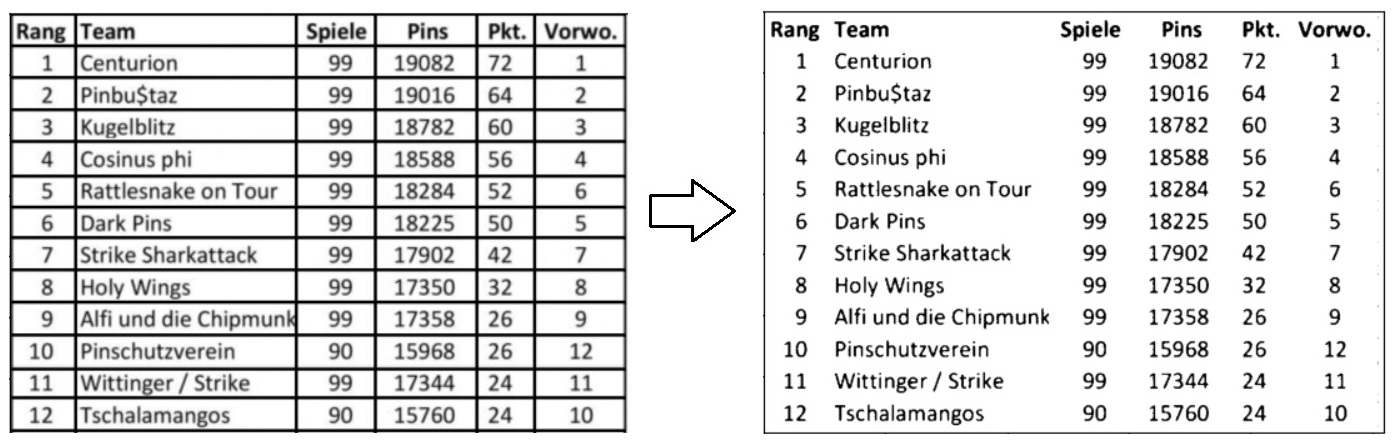}
  \caption{Example of the erosion operation as performed by Multi-Type-TD-TSR for partially bordered tables}
  \label{fig:erosion_partially_bordered}
\end{figure}
To handle all types of tables, an algorithm for partially bordered tables is needed. 
%
The main goal of our algorithms for bordered and unbordered tables is to create a grid cell image by adding borders in the unbordered case and detecting lines in the bordered case.
If a table is only partially bordered, then the unbordered algorithm is prevented to add borders in orthogonal direction to the existing borders, while the bordered algorithm can only find the existing borders. 
Both approaches result in incomplete grid cell images.
So the question is how to obtain an algorithm that produces a grid cell image for partially bordered tables. 
%
The main idea is to detect the existing borders as done by the algorithm for bordered tables, but without using them to create a grid cell, but to delete the borders from the table image to get an unbordered table
(see Figure \ref{fig:erosion_partially_bordered} for an example).
This allows then for applying the algorithm for unbordered tables to create the grid-cell image and contours by analogy to the variants discussed above.
A key feature of this approach is that it works with both bordered and unbordered tables: 
it is type-independent. 

\subsection{Color Invariance Pre-Processing}
\begin{figure}
  \centering
  \includegraphics[width=\linewidth]{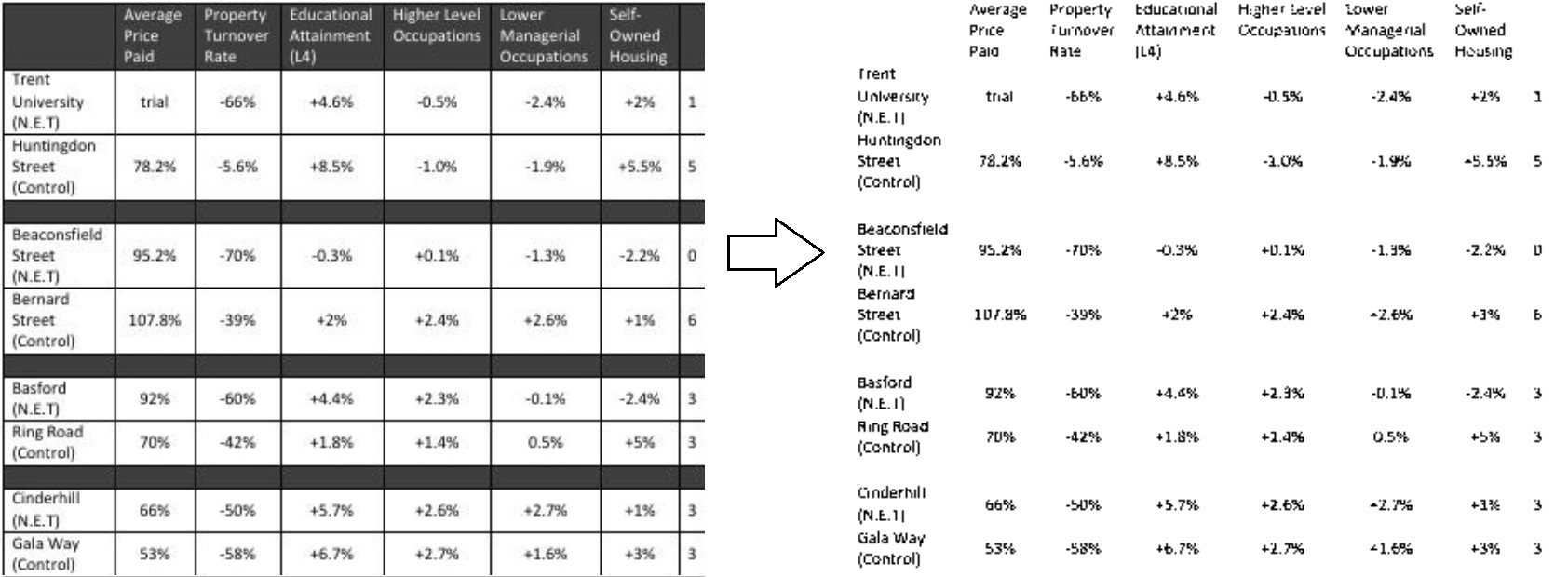}
  \caption{Color invariance pre-processing example}
  \label{fig:color_invariance}
\end{figure}

A main goal of this work is to create a multi-level pipeline for TD and TSR that works on all types of documents with tables. 
%
%
To this end, we addressed the problem of image rotation, detected tables in images, and developed an algorithm that can handle all three types of table margins. 
The quest is then whether this approach can handle different colors.
%
In general, we do not need to treat colors with 3 different channels as in the case of RGB images, for example, because we convert table images to binary images based on the contrast of font and background colors.
All algorithms proposed so far require a white background and a black font.
%
But the resulting binary image could have a black background and white font, or cells with a black background and white font, while others have a white background and black font as shown in Figure \ref{fig:color_invariance}.
Therefore, to obtain table images of the desired color type, we insert an additional image processing step between TD and TSR. 
This step also searches for contours, but now for counting black and white pixels per contour: 
if there are more black pixels than white pixels, the colors of the contour are inverted. 
This procedure results in backgrounds being white and fonts being black.  

\section{Evaluation}
\label{evaluation}
To enable comparability of Multi-Type-TD-TSR with other state-of-the-art approaches \cite{prasad2020cascadetabnet}, we reuse their datasets. 
%
This concerns a dataset for the TSR task that was extended by manually annotating selected images from the ICDAR 19 (Track B Modern) training data \cite{8978120}.
%
Prasad et al.\ \cite{prasad2020cascadetabnet} randomly chose 342 images out of 600 of the ICDAR 19 training set to get 534 tables and 24,920 cells, with all these entities annotated accordingly.
%
%
The reason for using only ICDAR 19 data is that the ground truth information available for the images as provided by the TableBank dataset \cite{li-etal-2020-tablebank} for TSR contains only table structure labels in the form of HTML tags.
%
It does not provide cell or column coordinates and therefore cannot be used to evaluate object detection performance. %
ICDAR 13 \cite{6628853} is also not usable for evaluating TSR because its evaluation metric uses the textual content of the cell-based mapping of predicted cells to ground truth ones.
This requires the extraction of text content with an OCR engine, so that the overall accuracy ultimately depends on the accuracy of the OCR. 

\begin{figure}
  \centering
  \includegraphics[width=\linewidth]{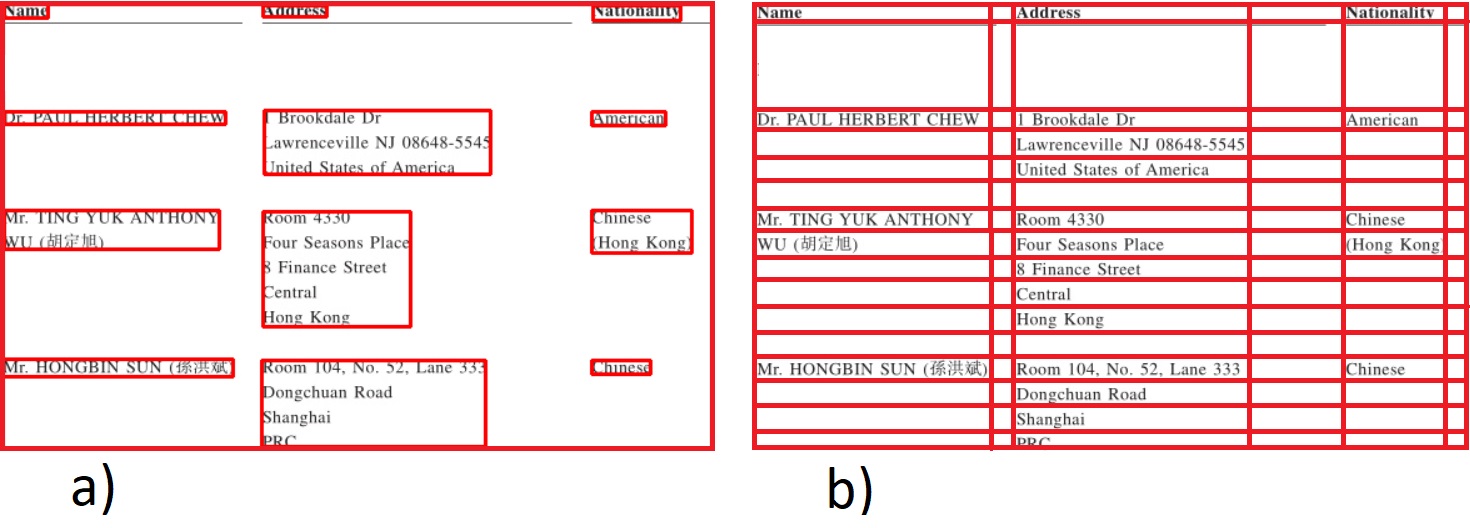}
  \caption{Annotation example from: a) the original validation dataset of \cite{prasad2020cascadetabnet}, b) the manually labeled validation dataset of Multi-Type-TD-TSR.}
  \label{fig:annotation}
\end{figure}

Our algorithm recognizes only cells as part of the overlap from recognized rows and columns.
%
To allow a fair comparison, we manually re-annotated the dataset with respect to the cells that our algorithm can recognize at all. 
An annotation example of both annotation types is shown in Figure \ref{fig:annotation}.

To validate our TSR algorithm, we need to determine the best kernel sizes for the horizontal and vertical kernel.
%
For this purpose, we used a random search to find the best values for the width of the vertical kernel and the height of the horizontal kernel.
%
We determined the best width to be 8 and the best height to be 3 pixel units. 

For the final evaluation, we used the type independent algorithm for partially bordered tables, since it is the one we would be deploying in a real world application where we do not have any information about the respective table types. 
%
We evaluate using F1-scores by analogy to \cite{prasad2020cascadetabnet} with IoU (\textit{Intersection over Union}) thresholds of 0.6, 0.7, 0.8, and 0.9, respectively. 
The results are shown in Table \ref{tab:table1}.
\begin{table}[h!]
    \centering
    \begin{tabular}{|c|c|c|c|c|c|}
    \hline
     \multirow{2}{*}{Team} & IoU & IoU & IoU & IoU & Weighted  \\
     & 0.6 & 0.7 & 0.8 & 0.9 & Average \\
     \hline
     CascadeTabNet & 0.438 & 0.354 & 0.19 & 0.036 & 0.232\\
     \hline
     NLPR-PAL & 0.365 & 0.305 & 0.195 & 0.035 & 0.206\\
     \hline
     Multi-Type- & \multirow{2}{*}{0.589} & \multirow{2}{*}{0.404} & \multirow{2}{*}{0.137} & \multirow{2}{*}{0.015} & \multirow{2}{*}{0.253} \\
     TD-TSR & & & & &\\
     \hline
\end{tabular}
    \caption{F1-score performances on ICDAR 19 Track B2 (Modern) \cite{8978120}}
    \label{tab:table1}
\end{table}
We achieved the highest F1-score by using a threshold of 0.6 and 0.7. 
When using higher thresholds (0.8 and 0.9), we encounter a clear performance decrease, which also applies for the other two algorithms we are comparing with.
According to the overall result, we conclude that Multi-Type-TD-TSR reaches the highest weighted average F1-score as well as the highest overall performance of 0.589, thus representing a new state-of-the-art.

\section{Conclusion}
\label{conclusion}
We presented a multistage pipeline for table detection and table structure recognition with document alignment and color invariance pre-processing. 
For this purpose, we  distinguished three types of tables, depending on whether they are borderless or not. 
Because of the unavailability of large labeled datasets for table structure recognition we decided to use 
%
two conventional algorithms: The first one can handle tables without borders, the second one can handle tables with borders. 
%
Further, we combined both algorithms into a third, conventional table structure recognition algorithm that can handle all three types of tables. 
%
This algorithm achieves the highest F1-score among the systems compared here for an IoU threshold of 0.6 and 0.7, but does not detect sharp borders, so the F1-score decreases rapidly for higher thresholds 0.8 and 0.9.
%
However, the highest weighted averaged F1-scores obtained by our algorithm show the potential of our multi-type approach, which can handle all three table types considered here:
it benefits from using one of the specialized algorithms to transform the input tables so that they can be optimally processed by the other specialized algorithm. 
This kind of multi-stage division of labor among otherwise conventional algorithms could help to finally bring such a difficult task as table structure recognition into domains that allow downstream NLP procedures to process linguistic table contents properly.
This paper made a contribution to this difficult task.

\section{Future Work}
\label{futurework}
The presented table structure recognition algorithms treat an approximation of the table structure recognition problem, because they assume that tables only contain cells as determined by the intersection of rows and columns. In general the problem of table structure recognition is even more difficult, since tables can recursively consist of more complex cells as found in tables with multi-rows or multi-columns. 
%
Cells can again consist of cells or contain entire tables, so that table structure recognition ultimately involves the recognition of recursive structures, which makes this task very difficult to handle for conventional computer vision algorithms.

The recent success of ML is due in part to the availability of large amounts of annotated data.
For table structure recognition such a dataset is not yet available so that many data driven algorithms use transfer learning to bypass this problem.
%
Obviously, larger datasets will lead to more general and better algorithms. Future work will therefore probably focus on the development of such datasets for table structure recognition.

\bibliographystyle{ACM-Reference-Format}
\bibliography{main}

\end{document}